\documentclass[conference]{IEEEtran}
\IEEEoverridecommandlockouts

\PassOptionsToPackage{dvipsnames}{xcolor}

\usepackage{cite}
\usepackage{amsmath,amssymb,amsfonts}
\usepackage{algorithmic}
\usepackage{graphicx}
\usepackage{url}
\usepackage{booktabs}
\usepackage{orcidlink}
\usepackage{multirow}
\usepackage{subfig}
\usepackage{textcomp}
\usepackage{xcolor}
\usepackage{hyperref}
\hypersetup{
    colorlinks=true,
    linkcolor={blue!100!black},
    citecolor={blue!100!black},
    urlcolor={blue!80!black}, 
    pdftitle={Predicting Surgical Safety Margins in Osteosarcoma Knee Resections: An Unsupervised Approach},
}

\def\BibTeX{{\rm B\kern-.05em{\sc i\kern-.025em b}\kern-.08em
    T\kern-.1667em\lower.7ex\hbox{E}\kern-.125emX}}
\begin{document}

\title{Predicting Surgical Safety Margins in Osteosarcoma Knee Resections: An Unsupervised Approach}

\author{\IEEEauthorblockN{Carolina Vargas-Ecos \orcidlink{0009-0001-6785-3652}}
\IEEEauthorblockA{\textit{Department of Biomedical Engineering} \\
\textit{Universidad Católica Boliviana ``San Pablo''}\\
La Paz, Bolivia \\
vargasecoscarolina@gmail.com}
\and
\IEEEauthorblockN{Edwin Salcedo \orcidlink{0000-0001-8970-8838}}
\IEEEauthorblockA{\textit{Department of Mechatronics Engineering} \\
\textit{Universidad Católica Boliviana ``San Pablo''}\\
La Paz, Bolivia \\
esalcedo@ucb.edu.bo}
}

\maketitle

\let\thefootnote\relax\footnote{\\978-1-6654-7811-3/25/\$31.00~\copyright2025 IEEE\hfill}

\begin{abstract}

According to the Pan American Health Organization, the number of cancer cases in Latin America was estimated at 4.2 million in 2022 and is projected to rise to 6.7 million by 2045. Osteosarcoma, one of the most common and deadly bone cancers affecting young people, is difficult to detect due to its unique texture and intensity. Surgical removal of osteosarcoma requires precise safety margins to ensure complete resection while preserving healthy tissue. Therefore, this study proposes a method for estimating the confidence interval of surgical safety margins in osteosarcoma surgery around the knee. The proposed approach uses MRI and X-ray data from open-source repositories, digital processing techniques, and unsupervised learning algorithms (such as k-means clustering) to define tumor boundaries. Experimental results highlight the potential for automated, patient-specific determination of safety margins. The full implementation and datasets are available at \url{https://github.com/EdwinTSalcedo/OsteosarcomaUCB}. 
\end{abstract}


\section{Introduction}

The term ``sarcoma'' encompasses a diverse group of malignant neoplasms, broadly categorized into bone sarcomas and soft tissue sarcomas. Osteosarcoma, the most prevalent primary bone cancer, most frequently arises in the metaphyseal region of long bones, particularly the distal femur, proximal tibia, and proximal humerus \cite{takeuchi2019}. It primarily affects children and adolescents between the ages of 10 and 20 \cite{ottaviani2010}, and is characterized by the excessive production of osteoid and compact bone tissue by malignant spindle cells \cite{samardziski2010}. Although rare, osteosarcomas are lethal, often presenting with few symptoms and in an advanced stage at diagnosis \cite{rothzerg2023}.

Treatment for osteosarcoma typically involves three stages: pre-chemotherapy, surgery, and post-chemotherapy \cite{majo2010}. Surgical intervention is crucial; it typically begins with an extended resection and progresses to a radical resection if necessary. An extended resection involves removing the biopsy site, the tumor, any potential metastases in the soft tissues adjacent to the skeletal tumor, and 2 to 5 centimeters of surrounding soft tissue that may contain micrometastases to achieve macroscopically clear margins. A radical resection, in contrast, entails the removal of entire muscle compartments, neurovascular bundles, or even the affected limb through amputation.

While limb-preserving surgery has become the standard, there is currently a lack of clear guidelines defining the ideal boundaries for removing a malignant osteosarcoma and determining how much healthy tissue must be excised to ensure clear margins and prevent recurrence \cite{takats2012}. Healthcare professionals often rely on the general surgical staging system for musculoskeletal sarcomas proposed by Enneking et al. \cite{enneking2003}, in which both extended and radical resections are classified based on criteria such as grade, site, and histogenic distribution. Specifically, the system categorizes musculoskeletal sarcomas as follows: IB – Low-Grade Extracompartmental, IIA – High-Grade Intracompartmental, and IIB – High-Grade Extracompartmental. These staging classifications can help avoid radical resections in some cases, and case-by-case feature analysis can aid in detecting highly aggressive sarcomas and potential metastasis to other organs.

\begin{figure*}
\centering
\includegraphics[width=\textwidth]{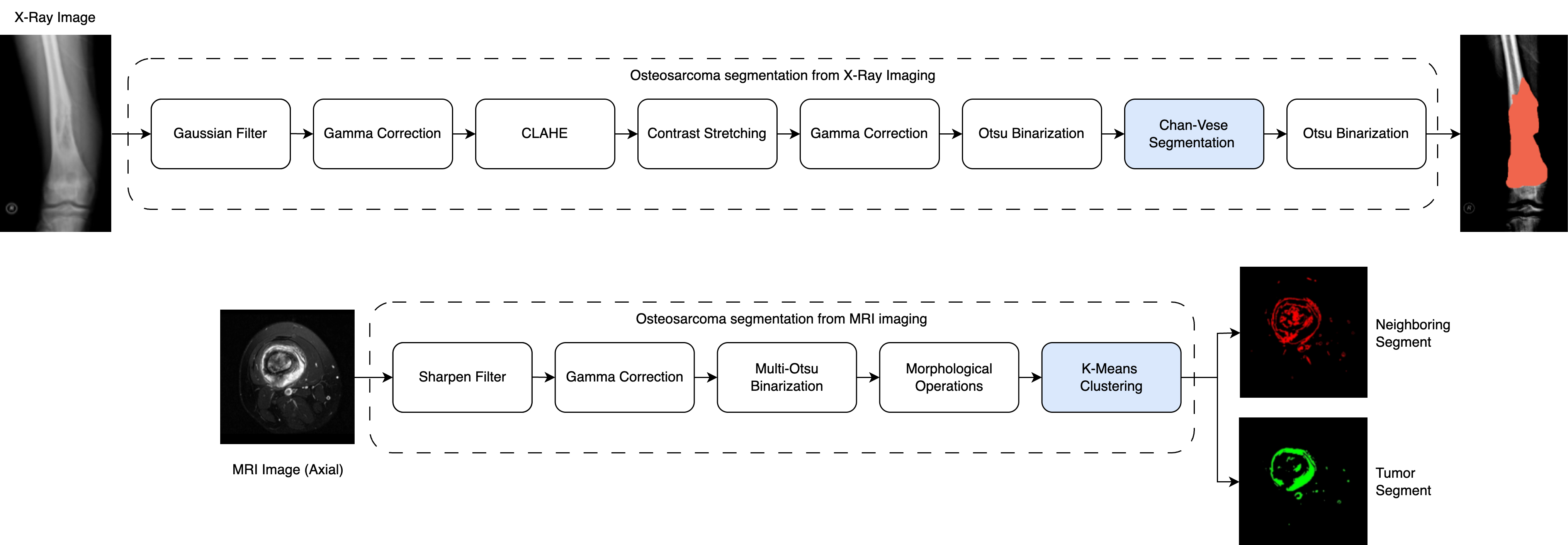}
\caption{Pipelines for X-Ray and MRI image segmentation.} \label{fig:pipeline}
\end{figure*}

Currently, the extent of healthy tissue removal in osteosarcoma surgery is typically determined by evaluating the patient’s medical history, imaging studies, and staging the tumor according to Enneking’s system \cite{enneking2003}. Yet, the definition of the safety margin for an extended resection remains somewhat empirical, often relying on a range of 3 to 5 centimeters around the affected area \cite{tinoco2021}. Recent progress in osteosarcoma segmentation using computer vision and deep learning from X-ray images \cite{von2021}, computed tomography (CT) scans \cite{shuai2023}, and magnetic resonance imaging (MRI) \cite{zhong2024} has shown precise results. However, most supervised approaches rely on large datasets that are either not publicly available or derived from populations in different regions, which may exhibit varying anatomical features. In response to this, Nasor et al. \cite{nasor2021} proposed an early approach to segment osteosarcoma from MRI images using unsupervised methods. Yet, like most other works in the literature, this approach only outputs segmented regions indicating tumor presence and does not define the margins of healthy tissue removal required for surgery.

Given the various complications that can arise from inadequate removal—such as the need for additional surgery due to tumor recurrence, as well as infection, fracture, limb loss, or even death \cite{henderson2014}—this study aims to investigate a method for estimating patient-specific surgical safety margins in knee osteosarcoma. Using open-source MRI and X-ray images, we apply image processing techniques and K-means clustering to define tumor boundaries and improve surgical precision, particularly in data-limited settings.

\section{Materials and Methods}

\begin{figure}
\centering
\includegraphics[width=0.5\textwidth]{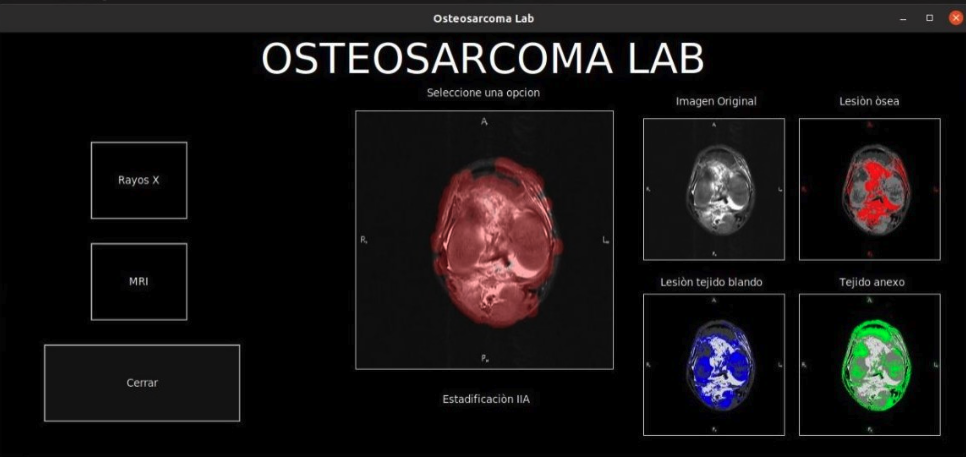}
\caption{GUI based on PyQT, which integrated the proposed algorithms to identify the safety margins.} \label{fig:gui}
\end{figure}

We propose the use of X-ray and MRI images to segment osteosarcoma in the knee and determine surgical safety margins. To this end, we developed two distinct segmentation pipelines (shown in Figure \ref{fig:pipeline}), designed to process new patient cases. To implement these pipelines, we compiled a new dataset from Radiopaedia.org \cite{radiopaedia2024} using the search term ``osteosarcoma.'' We then analyzed the resulting segments and manually measured the observed correlations to identify suitable surgical safety margins. Finally, we integrated these algorithms and margin calculations into a graphical user interface (GUI), shown in Figure \ref{fig:gui}, to facilitate their use.

\subsection{Data Collection and Exploration}

\begin{figure}[ht]
  \centering
  \subfloat[Patient age distribution.]{\includegraphics[width=0.48\textwidth]{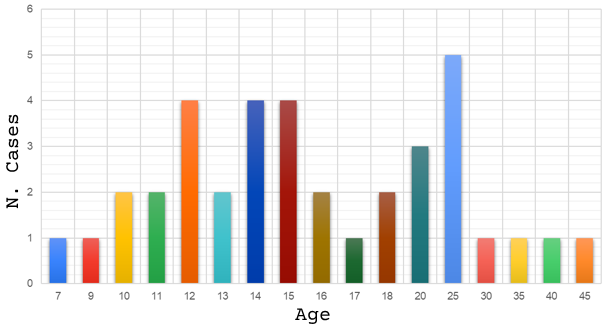}\label{fig:age-distribution}}
  \hfill
  \subfloat[Distribution of MRI images by anatomical plane.]{\includegraphics[width=0.40\textwidth]{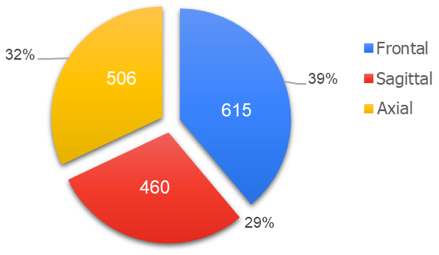}\label{fig:axis-distribution}}
  \caption{Demographics of the collected dataset.}
  \label{fig:Final end device}
\end{figure}

Radiopaedia.org \cite{radiopaedia2024} contains an open collection of articles, courses, and diagnosed medical cases with corresponding images. The collected data consisted of 54 patients with diagnostic images from X-ray, MRI, or both. Metadata for each image included the patient's origin, age, and gender. This resulted in an initial set of 1,399 images, of which 1,319 were MRI scans and 80 were radiographs. Figure \ref{fig:age-distribution} illustrates the age distribution within the dataset, showing the highest number of cases in the 10–20 age group. Regarding gender, 39 out of 54 reported cases were male. This is important, as gender directly influences bone length in data extrapolation. Additionally, the dataset revealed a predominance of osteosarcoma cases occurring around the knee.

As shown in Figure \ref{fig:axis-distribution}, the MRI samples in the dataset correspond to three anatomical planes: frontal (615), axial (460), and sagittal (506). Of the 54 total cases, 47 involved the femur, while the remaining 7 were located in the lower leg, specifically the tibia and fibula.

\subsection{Lesion Segmentation in X-Ray Images}
\label{sec:segm-rayx}

Osteosarcoma segmentation from X-ray imaging focuses on lesion identification and involves the following steps. First, we removed image noise using a Gaussian filter to distinguish bone structures from surrounding tissues. Then, gamma correction was applied to adjust the relationship between light input and output, enhancing bright structures and improving the visibility of lytic lesions. Subsequently, contrast-limited adaptive histogram equalization (CLAHE) was used to prevent noise amplification in homogeneous areas, such as joints. Next, the intensity range was adjusted using contrast stretching to create clearer density intervals. This step helped eliminate edema and soft tissue from the radiographs, prioritizing bone densities and identifying lytic lesions or ossified regions resulting from tumor growth that may be infiltrated by fluid. We then implemented Otsu binarization and Chan–Vese segmentation to identify the region of interest. Finally, a second gamma correction was applied to balance the overall contrast.

\subsection{Bone Tumor Segmentation on MRI}
\label{sec:segm-mri}

MRI segmentation enabled the clear delineation of lesioned areas from various tissue types, such as bone, muscle, and joints. The primary goal was to precisely define the boundaries of each distinct region. The implemented algorithm, shown in the second row of Figure \ref{fig:pipeline}, involves the following steps:

First, the visibility of fine anatomical structures (such as organ and tissue borders) was enhanced using a sharpening filter. Next, areas of varying intensities were balanced with gamma correction to prevent saturation in bright regions, allowing for easier differentiation between adipose, connective, and muscle tissue. Each image was then divided into multiple regions based on its histogram using multi-Otsu binarization, highlighting specific areas like tumors or lesions to minimize confusion between anatomical structures. Intermediate masks were refined using morphological operators. Finally, we applied K-means clustering to group the different tissues into three segments: background, tumor, and neighboring region.

\section{Experimental Results and Discussion}

This section evaluates both pipelines for segmenting osteosarcoma from X-ray and MRI images. Finally, we describe the analysis and measurement of surgical safety margins, leveraging the segmentation results.

\subsection{X-Ray Image Lesion Segmentation}

We evaluated the segmentation performance on X-ray images by comparing the masks generated by our algorithm (described in Section \ref{sec:segm-rayx}) with those created by an orthopedic surgeon. To this end, we calculated performance metrics including the Dice score, F1 score, sensitivity, accuracy, and Intersection over Union (IoU), as defined in \cite{salcedo2023}. The evaluation included images from 34 patients in the sagittal plane and 36 in the frontal plane. Table \ref{tab:results-sagittal} presents the detailed results of this analysis. All metrics were calculated on a per-pixel basis.

\begin{table}
\centering
\caption{Evaluation results for segmentation of frontal and sagittal radiographs.}\label{tab:results-sagittal}
\begin{tabular}{c|c|c}
\toprule
\textbf{Metric} &  \parbox{2.5cm}{\centering \textbf{Frontal X-ray \%}} &  \parbox{2.5cm}{\centering\textbf{Sagittal X-ray \%}} \\
\hline
Jaccard Micro Media &  92\% & 95\% \\
Media Accuracy Binary &  66\% & 70\% \\
Sensitivity &  79\% & 86\% \\
Media F1 Score Micro &  95\% & 97\% \\
Binary F1 Score Media &  78\% & 82\% \\
Media Recall Micro &  96\% & 97\% \\
Binary Media Recall &  78\% & 86\% \\
Media Dice Coefficient &  79\% & 82\% \\
\bottomrule
\end{tabular}
\end{table}

\begin{table}
\centering
\caption{Amount of data processed by the MRI segmentation algorithm.}\label{tab:results-mri}
\begin{tabular}{c|c|c}
\toprule
\textbf{MRI} &  \parbox{3cm}{\centering\textbf{Number of images processed}} &  \parbox{3cm}{\centering\textbf{Number of masks obtained}} \\
\hline
Frontal &  166 & 762 \\
Sagittal & 217 & 781 \\
Axial &  146 & 666 \\
Total &  529 & 2,209 \\
\bottomrule
\end{tabular}
\end{table}

\begin{figure}
\centering
\includegraphics[width=0.5\textwidth]{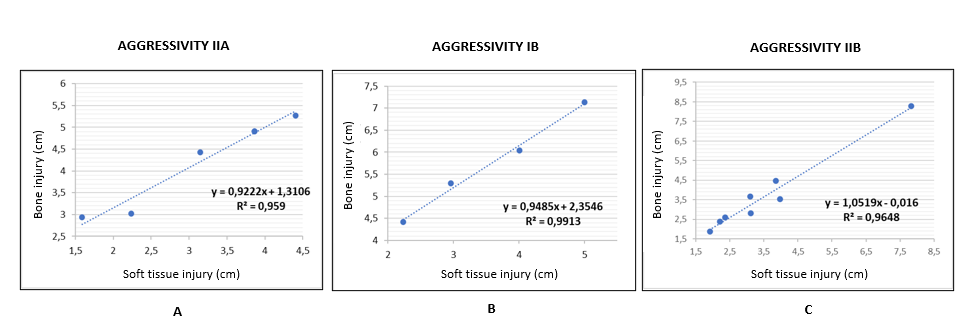}
\caption{Correlation between the value of the segmentation of the bone tumor lesion and the involvement of soft tissues with tumor lesion in Sagittal MRI.} \label{fig:correlation-mri}
\end{figure}

\subsection{Bone Tumor Segmentation on MRI and Prediction of Surgical Safety Margins}

For this imaging modality, we initially filtered out images with high saturation or low contrast, resulting in a total of 529 usable images. Subsequently, a total of 2,209 masks from frontal, sagittal, and axial MRI images were generated using the algorithm described in Section \ref{sec:segm-mri}. A summary of the data produced by the algorithm is presented in Table \ref{tab:results-mri}.

\begin{figure}[ht]
  \centering
  \subfloat[Measuring the length of the femur with the ImageJ program.]{\includegraphics[width=0.23\textwidth]{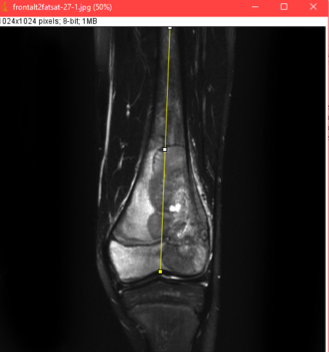}\label{fig:measuring-with-imagej}}
  \hfill
  \subfloat[Scaling pixels to cm femur length.]{\includegraphics[width=0.215\textwidth]{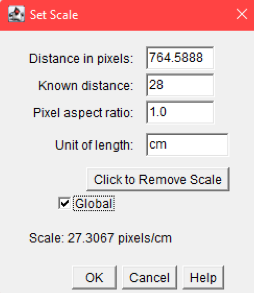}\label{fig:scaling-imagej}}
  \caption{ImageJ software used for annotating lengths.}
  \label{fig:imagj-software}
\end{figure}

Next, we investigated the correlation between segmentation values derived from the masks of 12 patients with sagittal images, 12 with frontal images, and 13 with axial images. The process began by calculating the mean of the extracted values for each mask, weighted by the number of pixels it contained. Figure \ref{fig:correlation-mri} displays the resulting trend line and correlation coefficient, specifically examining the relationship between bone tumor lesion mask values and soft tissue lesion mask values. This analysis indicates a connection between tumor growth and cell densification on one hand, and edema and muscle tissue damage on the other. Additionally, the summary suggests that this relationship is influenced by an aggressiveness index, which reflects both the speed of tumor growth and the duration that surrounding edema persists before it densifies. Furthermore, data from 18 patients were evaluated by calculating extramedullary tumor expansion into the muscle zone, with images individually scaled and measured using ImageJ. Figures \ref{fig:measuring-with-imagej} and \ref{fig:scaling-imagej} illustrate the measurement and scale-setting process, as applied to the case of an 18-year-old male patient. Femur length was estimated based on each patient’s sex and age to ensure accurate scaling.

Based on the observed correlations, surgical safety margins were determined at 0.25 cm intervals. Table \ref{tab:suggested-values} presents these values, with categories IIB, IB, and IIA considered according to Enneking’s staging system \cite{enneking2003} and assessed directly on the bone lesion due to their high aggressiveness and potential for new tumor growth. The analysis showed a highly linear relationship between bone lesion damage and adjacent soft tissue, with a data fit exceeding 92\%. This enables the prediction of tissue damage progression based on tumor size and staging. 

\begin{table}
\centering
\caption{Suggested values for surgical safety margin.}
\label{tab:suggested-values}
\begin{tabular}{c|c|c|c|c|c}
\toprule
\multicolumn{2}{c}{IIB} & \multicolumn{2}{c}{IB} & \multicolumn{2}{c}{IIA} \\
\hline
\parbox{1cm}{\centering Bone injury radius in [cm]} & \parbox{1.5cm}{\centering Surgical safety margin radius in [cm]} & \parbox{1cm}{\centering Soft tissue injury radius in [cm]} & \parbox{1cm}{\centering Surgical safety margin radius in [cm]} & \parbox{1cm}{\centering Soft tissue injury radius in [cm]} & \parbox{1cm}{\centering Surgical safety margin radius in [cm]} \\
\hline
0.50&	1.01990&	0.50&	2.828850&	0.50&	1.77170\\
0.75&	1.54585&	0.75&	3.065975&	0.75&	2.00225\\
1.00&	2.07180&	1.00&	3.303100&	1.00&	2.23280\\
1.25&	2.59775&	1.25&	3.540225&	1.25&	2.46335\\
1.50&	3.12370&	1.50&	3.777350&	1.50&	2.69390\\
1.75&	3.64965&	1.75&	4.014475&	1.75&	2.92445\\
2.00&	4.17560&	2.00&	4.251600&	2.00&	3.15500\\
2.25&	4.70155&	2.25&	4.488725&	2.25&	3.38555\\
2.50&	5.22750&	2.50&	4.725850&	2.50&	3.61610\\
2.75&	5.75345&	2.75&	4.962975&	2.75&	3.84665\\
3.00&	6.27940&	3.00&	5.200100&	3.00&	4.07720\\
3.25&	6.80535&	3.25&	5.437225&	3.25&	4.30775\\
3.50&	7.33130&	3.50&	5.674350&	3.50&	4.53830\\
3.75&	7.85725&	3.75&	5.911475&	3.75&	4.76885\\
4.00&	8.38320&	4.00&	6.148600&	4.00&	4.99940\\
4.25&	8.90915&	4.25&	6.385725&	4.25&	5.22995\\
4.50&	9.43510&	4.50&	6.622850&	4.50&	5.46050\\
4.75&	9.96105&	4.75&	6.859975&	4.75&	5.69105\\
\bottomrule
\end{tabular}
\end{table}

\section{Conclusions}

Automating the calculation of osteosarcoma resection safety margins is crucial, as current methods rely on empirically defined ranges. This study collected 1,399 medical images of osteosarcoma cases in the leg and knee regions to develop segmentation pipelines for distinguishing tumor, affected, and healthy tissue. Challenges included variability in imaging equipment, preprocessing requirements, and the need for patient anthropometric data to ensure accurate measurements. Axial and frontal images were prioritized for better damage assessment. We implemented a graphical user interface incorporating the proposed algorithms to assist professionals. Future work will expand the dataset to enable deep learning-based segmentation and explore additional unsupervised algorithms.

\bibliographystyle{ieeetr}
\bibliography{bibliography}

\end{document}